\documentclass[lettersize,journal]{IEEEtran}
\usepackage{amsmath,amsfonts}
\usepackage{algorithmic}
\usepackage{algorithm}
\usepackage{array}
\usepackage[caption=false,font=normalsize,labelfont=sf,textfont=sf]{subfig}
\usepackage{textcomp}
\usepackage{stfloats}
\usepackage{url}
\usepackage{verbatim}
\usepackage{graphicx}
\usepackage{cite}
\usepackage{amsmath}
\usepackage{amssymb}
\usepackage{booktabs}
\usepackage{multirow}
\usepackage{color}
\usepackage[dvipsnames]{xcolor}
\usepackage{colortbl}
\usepackage{amsfonts}
\usepackage{accents}
\usepackage{mathtools}
\usepackage{adjustbox}
\usepackage{bbding}
\usepackage{mathrsfs}
\usepackage{wrapfig}
\usepackage[colorlinks,linkcolor=red,anchorcolor=blue,citecolor=green]{hyperref}

\begin{document}

\title{MMF-Track: Multi-modal Multi-level Fusion for 3D Single Object Tracking}

\author{Zhiheng Li\dag, Yubo Cui\dag, Yu Lin, and Zheng Fang*,~\IEEEmembership{Member,~IEEE}
\thanks{\dag Authors with equal contribution.}
\thanks{The authors are all with the Faculty of Robot Science and Engineering, Northeastern University, Shenyang, China; Corresponding author: Zheng Fang, e-mail: fangzheng@mail.neu.edu.cn}
}

\markboth{Journal of \LaTeX\ Class Files,~Vol.~14, No.~8, August~2021}%
{Shell \MakeLowercase{\textit{et al.}}: A Sample Article Using IEEEtran.cls for IEEE Journals}

\maketitle

\begin{abstract}
3D single object tracking plays an important role in computer vision and autonomous driving. The mainstream methods mainly rely on point clouds to achieve geometry matching between target template and search area. However, textureless and incomplete point clouds make it difficult for single-modal trackers to distinguish objects with similar structures. 
To overcome the mentioned limitations of geometry matching, we propose a Multi-modal Multi-level Fusion Tracker (MMF-Track), 
which exploits the image texture and geometry characteristic of point clouds to track 3D target.
Specifically, we first propose a Space Alignment Module (SAM) to align RGB images with point clouds in 3D space, which is the prerequisite for constructing inter-modal associations.
After that, \textit{in feature interaction level}, we present a Feature Interaction Module (FIM) based on dual-stream structure, which enhances intra-modal features in parallel and constructs inter-modal semantic associations. Meanwhile, in order to refine each modal feature, we propose a Coarse-to-Fine Interaction Module (CFIM) to realize the hierarchical feature interaction at different scales.
Finally, \textit{in similarity fusion level}, we introduce a Similarity Fusion Module (SFM) to aggregate geometry and texture similarity from the target. 
Extensive experiments show that our method achieves competitive performance on KITTI and NuScenes datasets. The code will be opened soon in \url{https://github.com/LeoZhiheng/MMF-Tracker.git}.

\end{abstract}

\begin{IEEEkeywords}
3D single object tracking (SOT), Multi-modal data fusion, Transformer, Siamese network.
\end{IEEEkeywords}

\section{INTRODUCTION}
\label{sec:introduction}

\begin{figure}[t]
	\centering
	\includegraphics[width=9.0cm,height=8.4cm]{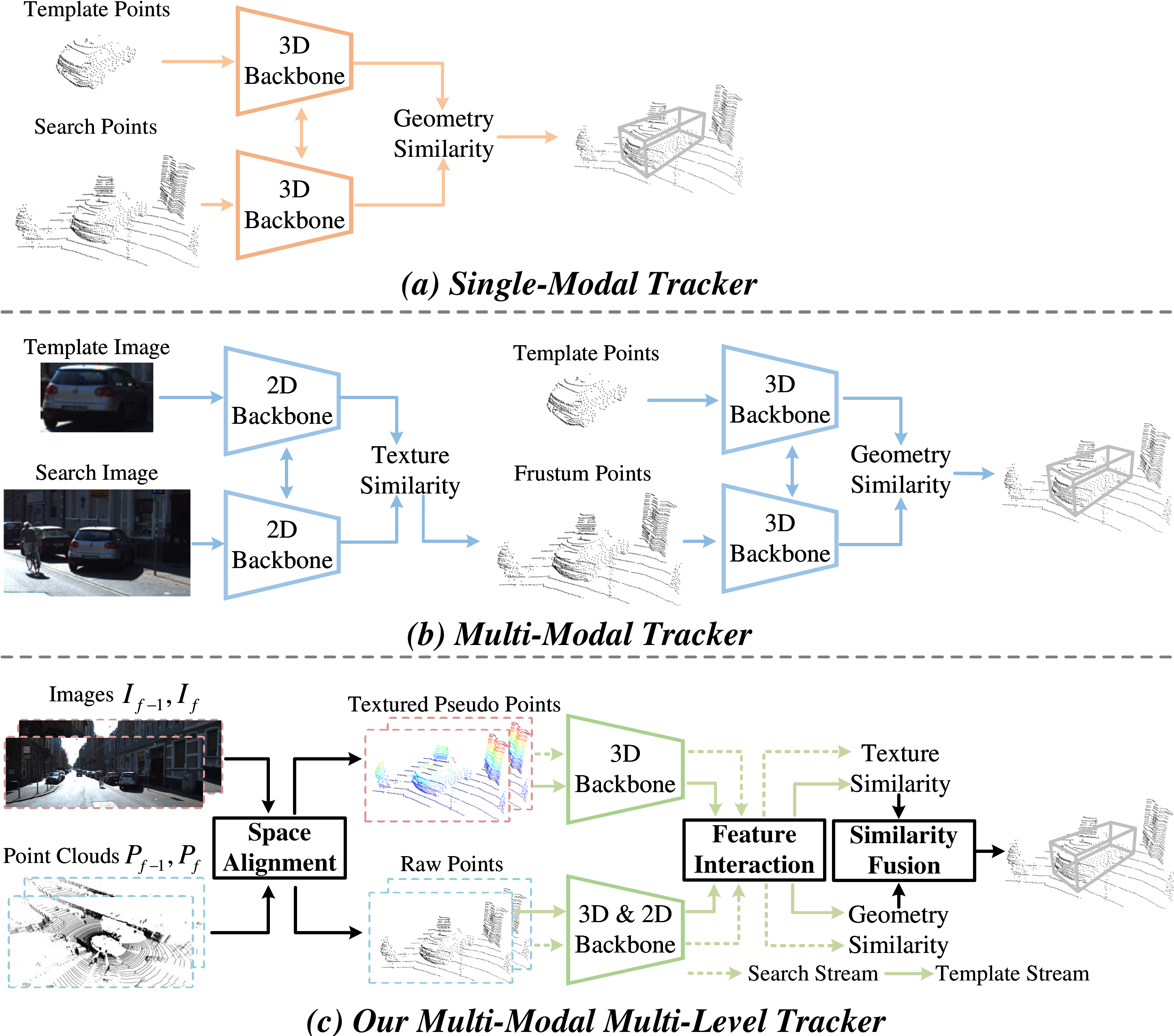}
	\caption{\textbf{Comparison of the current 3D single object tracking frameworks.} (a) Single-modal paradigm adopts the geometry matching of point clouds. (b) Multi-modal paradigm consists of 2D and 3D trackers in series. (c) Our MMF-Track employs dual-stream structure and includes three novel modules: space alignment, feature interaction and similarity fusion.}
	\label{fig:abstract}
\vspace{-0.08in}
\end{figure}

\IEEEPARstart{R}{ecently}, 3D single object tracking (3D SOT), as an important task of the computer vision, has attracted much attention in autonomous driving and mobile robots. Most 3D SOT methods~\cite{SC3D,P2B,BAT,PTT,lttr,v2b,pttr,DeepPCT} inherit 2D Siamese paradigm~\cite{SiamFC, siamDW, siamcar} and estimate the current target states via geometry matching between the target template and search region, as displayed in Fig.~\ref{fig:abstract}(a). However, due to these single-modal 3D trackers relying on sparse and textureless point clouds, it is difficult for them to identify target from the background with similar structures, such as tracking a specific person in the crowd. Fortunately, semantically rich images can provide fine-grained texture features, which could help tracker to distinguish target from disturbances.
Thus, it is valuable to explore multi-modal 3D SOT and combine the advantages of image and point cloud.

F-Siamese~\cite{F-siamese} is the first multi-modal SOT tracker, which adopts cascaded structure to connect the 2D tracker with 3D tracker, as shown in Fig.~\ref{fig:abstract}(b). Specifically, F-Siamese first adopts 2D tracker to estimate 2D bounding box of target and projects it into 3D viewing frustum based on camera matrix. Then, the 3D tracker extracts point features in the frustum and finally predicts the 3D target. 
However, F-Siamese overlooks the feature interaction between two modalities and loses many important cues from target. Thus, its performance lags much behind LiDAR-Only methods.

Apart from the cascaded tracker, the feature-level fusion is another effective way to aggregate multi-modal information, which has been verified in 3D object detection~\cite{pointpainting,pointaugmenting,clocs,epnet,deepfusion,autoalign}.
For example, PointPainting~\cite{pointpainting} and PointAugmenting~\cite{pointaugmenting} utilize segmentation scores or semantic features of images to decorate point clouds in the early stage, meanwhile CLOCs~\cite{clocs} merges cross-modal features at the instance level. Besides,~\cite{epnet,deepfusion,autoalign} deeply fuse hierarchical multi-modal features that are extracted from the 2D and 3D backbones. 
However, due to the different frameworks and objectives between tracking and detection, 3D SOT has its own challenges: (1) \textbf{Data Alignment}. Since tracking usually requires two branches as input, multi-modal tracker should align different sensor data and search-template areas, which is more complex and challenging than 3D detection task.
(2) \textbf{Feature Matching}. In order to locate the target within the search region, multi-modal tracker needs not only to focus on feature fusion but also consider effective feature matching between search and template branches.

To tackle the aforementioned issues, we propose a \textbf{M}ulti-modal \textbf{M}ulti-level \textbf{F}usion Tracker (MMF-Track), which aims to guarantee the data alignment for multi-sensor and integrate multi-modal information in the feature and similarity levels, as shown in Fig.~\ref{fig:abstract}(c).
Due to the heterogeneous representations between point cloud and image, it is non-trivial to effectively fuse 3D sparse points with 2D dense pixels. Thus, we first propose a \textit{Space Alignment Module} (SAM) to generate textured pseudo points aligned with raw points in 3D space. In this way, both of them can be further used to establish multi-modal semantic associations in the following network. 
Then, \textbf{in multi-modal feature interaction}, instead of simply combining two modal features, we propose a novel \textit{Feature Interaction Module} (FIM) based on dual-stream network, learning intra-modal long-range contextual information in parallel and then establishing inter-modal semantic associations through attention mechanism. 
Meanwhile, due to the limited semantic cues of single-scale feature, we further propose a \textit{Coarse-to-Fine Interaction Module} (CFIM), which exploits multi-scale texture information to refine two modalities by interpolation and encoding operation.
Moreover, \textbf{in geometry-texture similarity fusion}, different from previous methods that only utilize geometry similarity of raw points, we additionally generate texture similarity from pseudo points and adaptively fuse two similarities through \textit{Similarity Fusion Module} (SFM). Thus, the texture and geometry cues of target can be aggregated to search region, resulting in more robust tracking performance.

In summary, our contributions are as follows:
\begin{itemize}
\item We propose MMF-Track, a new multi-modal network for 3D SOT, which leverages both geometry and texture cues to track target.
\item We propose SAM to align different modalities and introduce CFIM to facilitate layer-wise feature interaction via a dual-stream structure. Additionally, we present SFM to aggregate the geometry and texture similarity of target.
\item Extensive experiments on KITTI and Nuscenes datasets
demonstrate that MMF-Track achieves promising performances, and the ablation studies prove the effectiveness of proposed modules. We will release the source code of our method to the community.
\end{itemize}

The rest of this paper is organized as follows. In Sec. \ref{sec:related-works}, we discuss the related work. Sec. \ref{sec:method} describes detailed structure of our MMF-Track. In Sec. \ref{sec:experiment} we first compare our methods with previous methods in KITTI and NuScenes datasets, and then conduct ablation studies to explore effectiveness of each module in our methods. Finally, we conclude in Sec. \ref{sec:conclusion}.

\section{RELATED WORK}
\label{sec:related-works}

\subsection{2D Single Object Tracking} 

Early works in visual SOT usually utilize correlation filters~\cite{Staple, ECO, tim_3} to track object. However, due to depending on template matching and seldom using deep features from CNN, they usually fail to capture fast-moving objects. Recently, the Siamese-like pipeline~\cite{GOTURN, SiamFC, SiamRPN} has received widespread attention in the visual SOT community. GOTURN~\cite{GOTURN} is the pioneering work which fed the template and search frames to the two branches of the Siamese framework and extracted features. Then, GOTURN~\cite{GOTURN} concatenated the two branch features to regress target location. Inspired by~\cite{GOTURN}, SiamFC~\cite{SiamFC} exploited Siamese network to compute the correlation feature and utilized it to locate target position in the search image. However, since they only employed the network to classify, the precision of predicted 2D bounding box is not good enough. Thus, SiamRPN~\cite{SiamRPN} introduced the idea of region proposal network (RPN) that consisted of two heads for classification and regression respectively, resulting in higher regression precision. Besides, SiamCAR~\cite{siamcar} and SiamBAN~\cite{siameseban} proposed a pixel-wise regression method to avoid generating numerous anchors. DiMP~\cite{DiMP} employed a meta-learning formulation to obtain a robust classification model for target and background. ATOM~\cite{ATOM} utilized overlap maximization and conjugate gradients to achieve accurate predictions. Moreover, many recent works~\cite{transformermeetstracker,transformertracking,tim_2} have exploited Transformer~\cite{Transformer} to achieve significant performance improvement. Although 2D trackers have obtained promising results in the large-scale datasets and practical applications, they are still difficult to handle 3D point clouds for 3D SOT.

\subsection{3D Single Object Tracking}
Inspired by 2D SOT, the previous 3D SOT approaches~\cite{P2B, BAT, v2b} usually calculate the point-wise similarity between the template and search branches. Specifically, P2B~\cite{P2B} further embedded cosine similarity map with target clues into search feature and predicted the target through VoteNet~\cite{votenet}. Later, BAT~\cite{BAT} proposed BoxCloud to encode the prior knowledge of target dimension to strengthen the similarity learning ability.
Furthermore, V2B~\cite{v2b} voxelized discrete points to generate the dense BEV features and regressed 3D target in a pixel-wise manner.
With the popularity of Transformer~\cite{Transformer} in computer vision, some works~\cite{PTT, lttr, pttr, stnet, SMAT} attempted to apply Transformer to 3D SOT and obtained stronger tracking performance. Specifically, PTT~\cite{PTT} adopted attention mechanism to weigh point cloud features. LTTR~\cite{lttr} exploited Transformer to capture the inter- and intra- associations among the different regions. Additionally, PTTR~\cite{pttr}, STNet~\cite{stnet} and SMAT~\cite{SMAT} proposed Transformer encoder-decoder to enhance and merge features. Different from the above similarity approaches, M$^2$-Track~\cite{beyond} proposed a motion-centric paradigm and estimated target motion state through point clouds of consecutive frames. \cite{beyond} had greatly overcome the problem of point cloud sparsity and achieved significant performance improvement. 

Nevertheless, no matter the motion or similarity paradigm, these methods relied on sparse point clouds, leading trackers to easily confuse the objects with similar geometry. To the best of our knowledge, there are few works exploring the multi-modal 3D SOT task.
F-Siamese~\cite{F-siamese} is the most recent multi-modal tracker, which utilized 2D tracking result to generate the 3D frustum and adopted 3D tracker to predict target. However, its cascaded structure limits the potential capabilities of multi-modal tracker.

\begin{figure*}[ht]
\centering
\includegraphics[width=\linewidth,height=6.2cm]{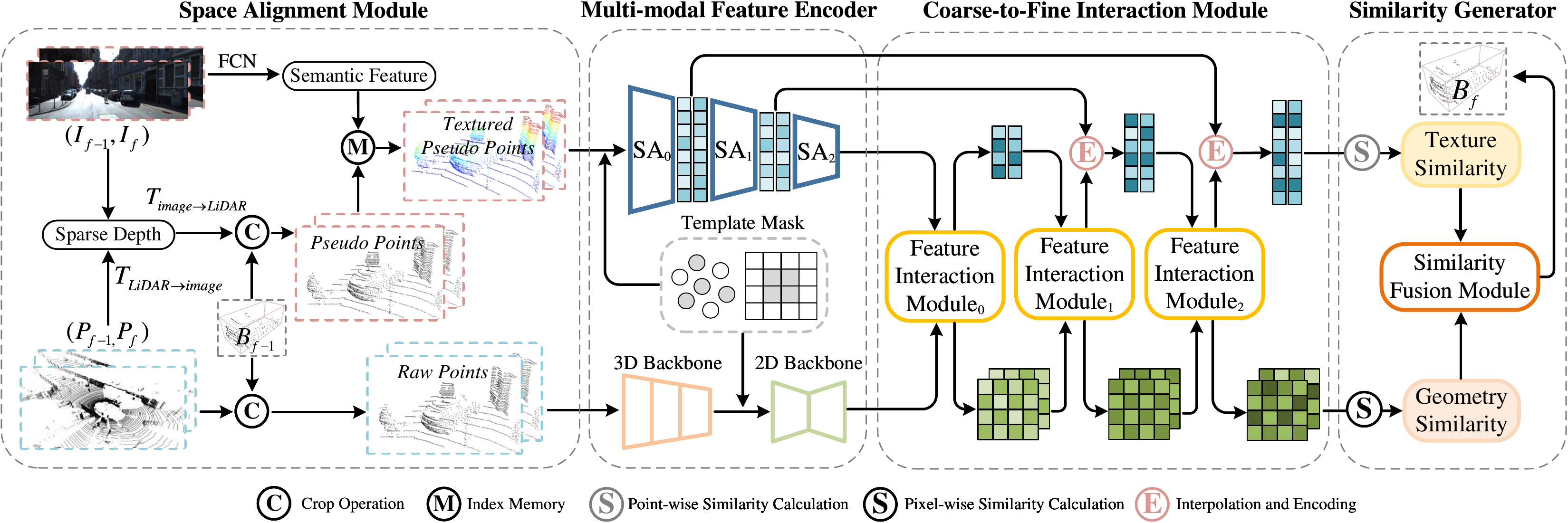}
\caption{\textbf{Overview of our proposed MMF-Track.} The network consists of four steps: (a) \textbf{\textit{Space Alignment Module}} generates textured pseudo points parallel to the raw points branch. (b) Multi-modal Feature Encoder extracts discriminative features from two modalities. (c) \textbf{\textit{Coarse-to-Fine Interaction Module}} adopts \textbf{\textit{Feature Interaction Module}} and interpolation operation to enhance intra-modal features and construct inter-modal associations iteratively. (d) Similarity Generator generates geometry and texture similarities and then fuses them via \textbf{\textit{Similarity Fusion Module}} for regressing the target.}
\label{pic: network overview}
\end{figure*}

\subsection{Multi-modal Feature Fusion in 3D Object Detection}
Though there are few multi-modal 3D SOT methods, many 3D detection methods combining the advantage of image with point clouds through feature fusion have been proposed to overcome the limitations of separate sensors. These methods can be divided into three categories: \textit{early fusion}~\cite{pointpainting,pointaugmenting}, \textit{late fusion}~\cite{clocs} and \textit{deep fusion}~\cite{epnet,deepfusion,autoalign}. 
For example, PointPainting~\cite{pointpainting} decorated image segmentation scores for each point and generated painted points as network input. Different from \cite{pointpainting}, CLOCs~\cite{clocs} generated 2D and 3D proposals, which were merged at the object level. Besides, most approaches (e.g. EPNet~\cite{epnet}) deeply combined features extracted from different backbones and made full use of semantic information in multi-modal. Inspired by aforementioned works, we design a feature interaction network that is more suitable for tracking task and propose a novel similarity fusion method to aggregate texture and geometry clues of target.

\section{METHODOLOGY}
\label{sec:method}

\subsection{Overview}
Given initial bounding box $B_0=(x_0,y_0,z_0,w,h,l,\theta_0)\in R^{7}$ of the specific target at the first frame, 3D single object tracking aims to utilize sensor data to predict the target state $B_f=(x_f,y_f,z_f,\theta_f) \in R^{4},f\in\{1,...,m\}$ frame by frame, where $(x, y, z)$ is the bounding box center, and $\theta$ is the heading angle. Following the previous works~\cite{P2B,BAT,PTT}, we assume that the target size $(w, h, l)$ is unchanged during tracking.

According to the number of modalities employed, trackers can be divided into single-modal and multi-modal. And the former can be formulated as follows:
\begin{equation}
\Psi(B_{f-1},P_{f-1},P_{f}) \mapsto B_{f}
\end{equation}
where $P_{f-1},P_{f}$ are the two consecutive point clouds. Furthermore, the multi-modal tracker~\cite{F-siamese} adds image tracking phase before 3D tracker, forming a two-stage tracking paradigm:
\begin{gather}
\Omega(\hat{B}_{f-1},I_{f-1},I_{f}) \mapsto \hat{B}_{f} \\
\Phi({B}_{f-1},\hat{B}_{f},P_{f-1},P_{f}) \mapsto B_{f}
\end{gather}
where $(I_{f-1},I_{f}), (\hat{B}_{f-1},\hat{B}_{f})$ represent images and predicted 2D bounding boxes in the last and current frames respectively.
Different from the above paradigms, we design a dual-stream network that processes multi-modal features in parallel, which is shown in Fig.~\ref{pic: network overview} and formulated as follows:
\begin{equation}
\Gamma(B_{f-1},I_{f-1},I_{f},P_{f-1},P_{f}) \mapsto {B}_{f}
\end{equation}

\subsection{Data Alignment and Feature Extraction}
\subsubsection{Space Alignment Module}~\label{sec:SAM}
Due to the gap in data format between images and point clouds, we propose a Space Alignment Module (SAM) to ensure multi-modal spatial alignment. Specifically, we first project the point clouds $P\in\{P_{f-1},P_{f}\}$ to the image plane through projection matrix $T_{LiDAR\rightarrow image}$ and allocate the depth of each point to its nearest pixel $I_{ij}$.
Thus, we can generate sparse depth maps $S\in\{S_{f-1},S_{f}\}$ with the size of $H\times W\times 1$ and convert $S$ into pseudo points by matrix $T_{image\rightarrow LiDAR}$, where $W$ and $H$ are the width and height of image.
When given predicted box $B_{f-1}$ in the previous frame, we crop out multi-modal search-template points and store indexes that indicate the projection relationship between the pixels and pseudo points. Inspired by \cite{pointaugmenting}, we further adopt FCN~\cite{fcn} to extract semantic features $Y\in\{Y_{f-1},Y_{f}\}$ from images $I\in\{I_{f-1},I_{f}\}$ and append $Y$ to pseudo points by the index memory. Finally, SAM output textured pseudo points $(P_{p}^{t}, P_{p}^{s})$ and raw points $(P_{r}^{t}, P_{r}^{s})$.

Note that our SAM has the following novelties: 1) Different from \cite{pointaugmenting} using image features to augment the raw points, we generate textured pseudo
points parallel to raw points branch, which is convenient for subsequent multi-modal semantic association. Besides, FCN~\cite{fcn} can be trained with our network and online decorate points rather than offline painting features as in \cite{pointaugmenting}. 2) Instead of viewing the point clouds in $B_{f-1}$ as template, we argue that contextual information of background could help tracker perceive the target motion. Therefore, we crop out the template with the same size as search region and maintain background points surrounding target. 

\begin{figure}[t]
\centering
\vspace{0.05in}
\includegraphics[width=8.2cm,height=8.5cm]{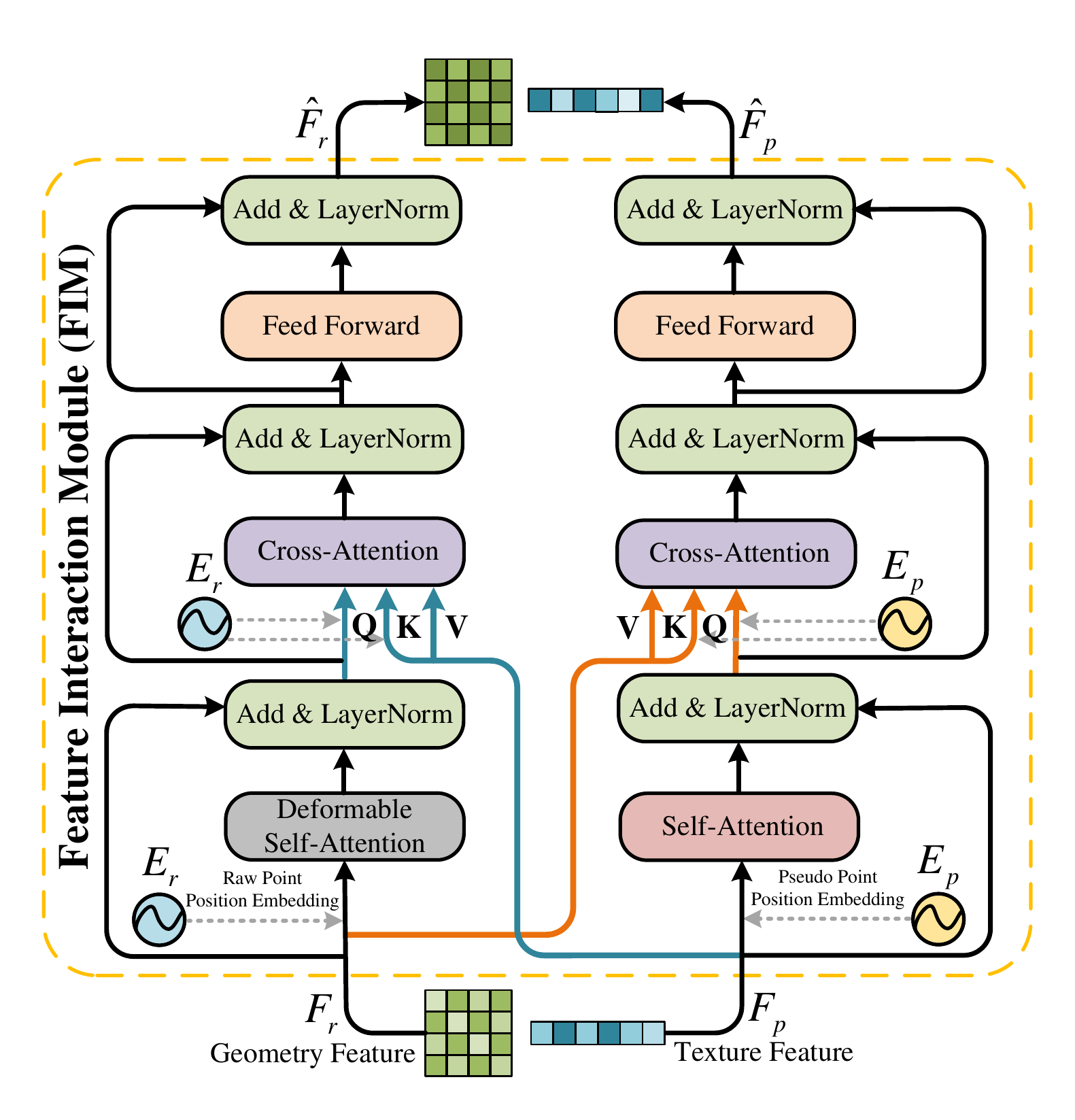}
\caption{\textbf{Dual-stream structure of Feature Interaction
Module (FIM).} Self-attention and cross-attention are utilized to enhance two modal features and construct semantic associations.}
\label{pic:FIM}
\vspace{-0.1in}
\end{figure}

\subsubsection{Multi-modal Feature Encoder}
To avoid confusion between target and background information in template branch, we introduce template mask $M$ to distinguish target and background points. Specially, for the textured pseudo points, the mask $M_{p_i} (i\in\{1,2,...,N\})$ indicates whether the template point $P_{p_{i}}^{t}$ is located in $B_{f-1}$ by padding with 1 or 0, where $N$ is the number of points. After combining points $P_{p}^{t}$ with mask $M_p\in R^{N\times 1}$, we feed pseudo points $P_p\in \{P_{p}^{t}\odot M_{p}, P_{p}^{s}\}$ into PointNet++~\cite{PointNet++} to generate texture features $F_{p}\in \{F_{p}^{t},F_{p}^{s}\}$:
\begin{equation}
F_{p}^{t}=\text{PointNet++}_{1}(P_{p}^{t}\odot M_{p}),
F_{p}^{s}=\text{PointNet++}_{2}(P_{p}^{s})
\end{equation}
where $\odot$ is feature concatenation along channel dimension. 
For the raw points, we first utilize 3D backbone of SECOND~\cite{second} to generate dense BEV features $G_{r}\in \{G_{r}^{t},G_{r}^{s}\}$ with the size of $H_1\times W_1\times D$ and merge $G_{r}^t$ with $M_r$, where $W_1$ and $H_1$ denotes the width and height of BEV features. Note that $M_{r}\in R^{H_1\times W_1\times 1}$ is similar to $M_{p}$ but projected to the BEV plane. Then, we adopt a group of convolution as 2D Backbone to further encode $G_{r}$ and get geometry features $F_{r}\in \{F_{r}^{t},F_{r}^{s}\}$ with the same size as $G_{r}$:
\begin{equation}
F_{r}^{t}=\text{Conv}_{1}(G_{r}^{t}\odot M_{r}),
F_{r}^{s}=\text{Conv}_{2}(G_{r}^{s})
\end{equation}

\subsection{Feature Enhancement and Interaction}~\label{sec:FIM}
In this part, we design a dual-stream structure to propagate multi-modal information bidirectionally and propose a coarse-to-fine strategy to refine each modal iteratively.
To simplify description, we explain the search branch that is the same as the template and replace $(F_{p}^{s}, F_{r}^{s})$ with $(F_p, F_r)$.
\subsubsection{Feature Interaction Module} 
To enhance multi-modal features which represent the same semantic object, we propose a Feature Interaction Module (FIM) composed of intra-modal feature enhancement and inter-modal semantic interaction.

\noindent
\textbf{Intra-modal Feature Enhancement.} In Fig.~\ref{pic:FIM}, we exploit vanilla self-attention to learn long-range contextual information in ${F}_{p}$ and obtain a more discriminative texture feature $\mathcal{{F}}_{p}$ by attention weights. 
Meanwhile, to reduce computational complexity of attention mechanism, we employ deformable attention \cite{DeformableDETR} to $F_{r}$ and get a non-local geometry feature $\mathcal{{F}}_{r}$. The process of feature enhancement is described as follows:
\begin{gather}\label{eq:QKV}
Q=\alpha(F)+E,\, K=\beta(F)+E,\, V=\gamma(F) \\
\mathcal{{F}}_{p}=\text{LN}(\text{SelfAttn}(Q_p,K_p,V_p))+Q_p\\
\mathcal{{F}}_{r}=\text{LN}(\text{DeformAttn}(Q_r,K_r,V_r))+Q_r
\end{gather}
where $E\in \{E_{p}, E_{r}\}$ is the position encoding of two modalities. LN means layer normalization, and $\alpha,\beta,\gamma$ represent three linear layers with non-shared weights.

\noindent
\textbf{Inter-modal Feature Interaction.} 
Due to the two modalities focusing on different information, we design a dual-stream structure to realize bilateral interaction rather than the classical unilateral fusion.
To construct a semantic association between the texture and geometry features, we first convert $\mathcal{{F}}_{r}$ to $\mathcal{{Q}}_{r}$ by Eq.~\ref{eq:QKV} in the geometry branch and map ${{F}}_{p}$ to $(\mathcal{{K}}_{p}, \mathcal{{V}}_{p})$ in another branch. After that, we feed $(\mathcal{{Q}}_{r}, \mathcal{{K}}_{p}, \mathcal{{V}}_{p})$ to cross-attention operation, which encodes semantic consistent texture information to geometry feature and generate a more robust feature representation $\mathscr{{F}}_{r}$. Meanwhile, in the texture branch, we also utilize the same approach to enhance the semantic information of texture feature $\mathcal{{F}}_{p}$:
\begin{gather}
\mathscr{{F}}_{p}=\text{LN}(\text{CrossAttn}(\mathcal{{Q}}_{p},\mathcal{{K}}_{r},\mathcal{{V}}_{r}))+\mathcal{{Q}}_{p}\\
\mathscr{{F}}_{r}=\text{LN}(\text{CrossAttn}(\mathcal{{Q}}_{r},\mathcal{{K}}_{p},\mathcal{V}_{p}))+\mathcal{{Q}}_{r}
\end{gather}

\begin{figure}[t]
\centering
\includegraphics[width=\linewidth,height=7.5cm]{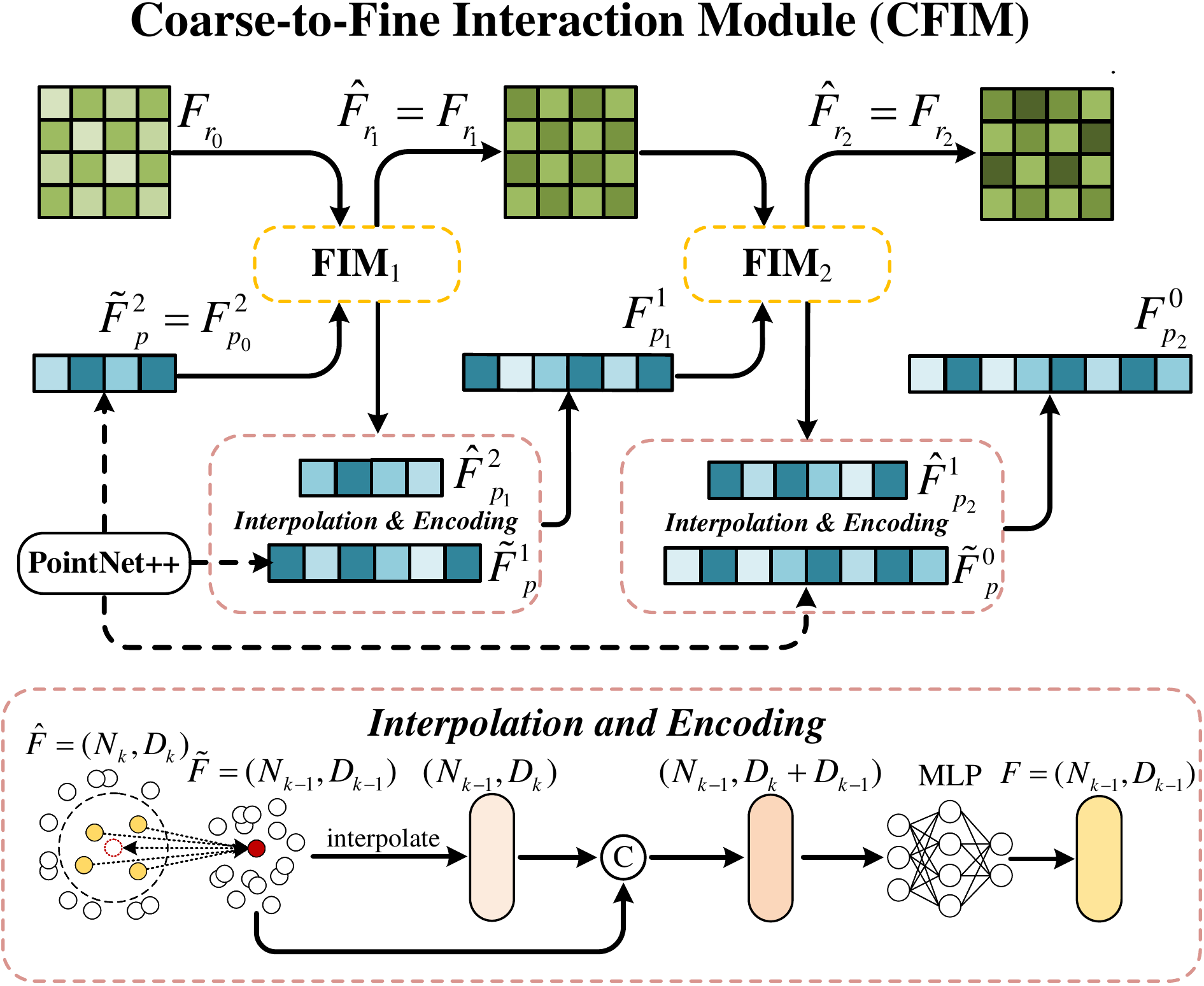}
\caption{\textbf{The detail of Coarse-to-Fine Interaction Module (CFIM).} CFIM utilizes multi-scale information to refine multi-modal features iteratively by Interpolation and Encoding operation as well as Feature Interaction Module.}
\label{pic:CFIM}
\vspace{-0.1in}
\end{figure}

Finally, we feed $(\mathscr{F}_{p}, \mathscr{F}_{r})$ to fully connected feed-forward network and generate enhanced features $(\hat{F}_p, \hat{F}_r)$. After the above steps, we not only realize multi-modal interaction but also retain each modal feature independently. Thus the features can be refined iteratively and generate geometry and texture similarities in the following modules.

\subsubsection{Coarse-to-Fine Interaction Module}\label{section:sec3.3.2}
Although simply stacking FIM could enhance two modal features, it is still difficult to construct complex semantic associations depending on single-scale features. 
Meanwhile, due to the down-sampling operation in the feature encoder, the number of texture features becomes very small at the end, limiting multi-modal interaction in FIM.
Therefore, we propose a Coarse-to-Fine Interaction Module (CFIM) that exploits multi-scale features to achieve multi-modal interaction layer by layer and finally obtains finer features. 
As shown in Fig.~\ref{pic:CFIM}, we first utilize layer-wise texture features $\tilde{F}_{p}^{k}\in R^{N_k\times D_k}(k\in\{0,1,2\})$ extracted by PointNet++~\cite{PointNet++} as the source of multi-scale information, where $N_k$ and $D_k$ mean the seed point number and channel dimension in the $k$-th set-abstraction layer. Then, in the $j$-th CFIM layer $(j\in\{1,2\})$, we apply FIM$_{j}$ to interact $F_{p_{j-1}}^{k}$ with $F_{r_{j-1}}$ and generate $\hat{F}_{p_{j}}^{k}\in R^{N_{k}\times D_{k}}, \hat{F}_{r_{j}}\in R^{
H_1\times W_1\times D_{k}}$.
\begin{equation}
\hat{F}_{p_{j}}^{k}, \hat{F}_{r_{j}} = \text{FIM$_{j}$}(F_{p_{j-1}}^{k}, F_{r_{j-1}})
\end{equation}

Later, following feature propagation in \cite{PointNet++}, we interpolate $\hat{F}_{p_{j}}^{k}$ with $\tilde{F}_{p}^{k-1}$ and adopt a MLP to generate feature ${F}_{p_{j}}^{k-1}\in R^{N_{k-1}\times D_{k-1}}$ by Eq.~\ref{eq:interpolate}, where $\tau$ is the interpolation function.
\begin{equation}\label{eq:interpolate}
{F}_{p_{j}}^{k-1}=\text{MLP}(\text{Concat}(\tau(\hat{F}_{p_{j}}^{k}, \tilde{F}_{p}^{k-1}), \tilde{F}_{p}^{k-1}))
\end{equation}

Finally, by repeating the above processes, our CFIM could utilize multi-scale information to complement and refine multi-modal features iteratively.

\begin{figure}[t]
\centering
\includegraphics[width=\linewidth,height=7.6cm]{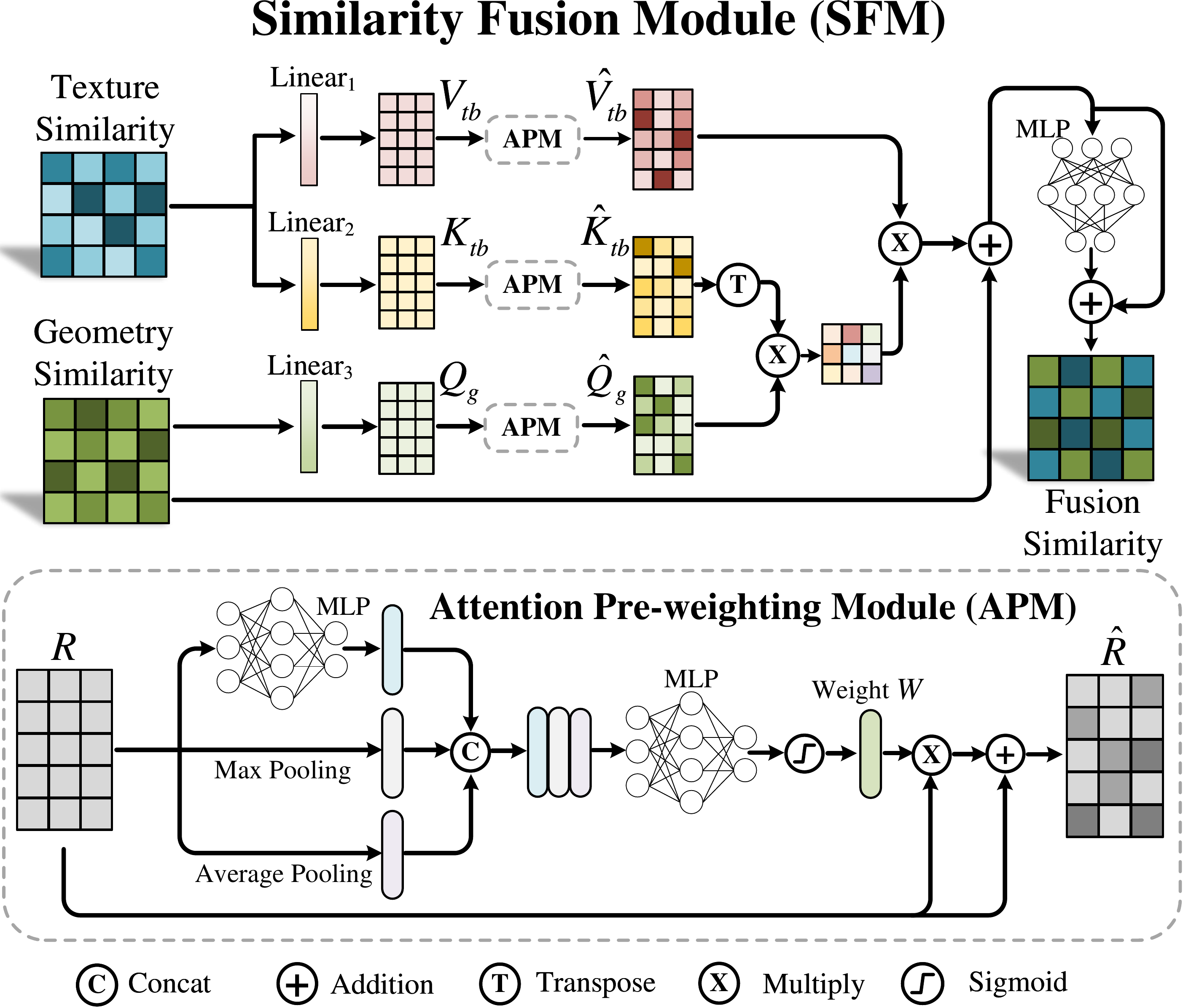}
\caption{\textbf{The structure of Similarity Fusion Module (SFM).} SFM enhances each-modal similarity through Attention Pre-weighting Module (APM) before adaptively fusing geometry and texture similarities.} 
\label{pic:SFM}
\vspace{-0.1in}
\end{figure}

\subsection{Similarity Generation and Fusion}\label{section:sec3.4}

\noindent
\textbf{Similarity Generation.} For geometry features $(F_{r}^{t},F_{r}^{s})$ from CFIM, we calculate the pixel-wise similarity~\cite{Alpharefine} and generate geometry similarity $S_{g}\in R^{H_1\times W_1 \times D}$. Meanwhile, for texture feature $(F_{p}^{t},F_{p}^{s})$, we use cosine distance as metric and generate point-wise texture similarity $S_{t}\in R^{N_{0}\times D}$. Then, to ensure feature alignment, we perform the pillarization process by Max Pooling to convert $S_{t}$ into BEV feature $S_{tb}\in R^{H_1\times W_1 \times D}$.

\noindent
\textbf{Similarity Fusion.} We propose a Similarity Fusion Module (SFM) that adaptively combines $S_{g}$ with $S_{tb}$ and aggregates geometry and texture clues from target. Specifically, as shown in Fig.~\ref{pic:SFM}, we first use three linear layers to generate attention vector $Q_{g}, K_{tb}, V_{tb}$, respectively. Then, to maximize the use of significant information in each-modal similarity, we propose an Attention Pre-weighting Module (APM) to enhance target clues in attention vector. The APM is described as follows:
\begin{gather}
W = \sigma(\text{MLP}(\text{Concat}(\text{MLP}(R),\text{Max}(R),\text{Avg}(R))))\\
\hat{R} = RW + R
\end{gather}
where $\sigma$ is the sigmoid function and $R \in \{{Q}_{g},{K}_{tb},{V}_{tb}\}$, $\hat{R} \in \{\hat{Q}_{g},\hat{K}_{tb},\hat{V}_{tb}\}$. $\text{Max}$ and $\text{Avg}$ represent Max Pooling and Average Pooling, respectively.
Finally, we adopt the pre-weighted attention vector $\hat{Q}_{g},\hat{K}_{tb},\hat{V}_{tb}$ to calculate the attention weight and achieve similarity fusion by Eq.~\ref{eq:SFM}.
\begin{gather}\label{eq:SFM}
\hat{S} = \frac{\hat{Q}_{g}{\hat{K}_{tb}}^T}{\sqrt{d_k}}\hat{V}_{tb} + S_{g},\, S = \text{MLP}(\hat{S}) + \hat{S}
\end{gather}
where $\sqrt{d_k}$ is the scale factor. Different from vanilla cross-attention, our proposed SFM adaptively enhances the critical target clues in geometry and texture similarities before exchanging information with each other. Thus, compared with simply stacking, adding, or directly using cross-attention, SFM can obtain a more discriminative similarity feature $S$.

\subsection{Prediction and Training}
Following CenterPoint~\cite{centerpoint}, we use fusion similarity $S$ to predict the target's spatial location and orientation in a center manner. The training loss is formulated as follows:

\begin{equation}
    L = \lambda_{cls}L_{cls}+\lambda_{reg}L_{reg}
\end{equation}
where we use focal loss~\cite{focalloss} as $L_{cls}$ to train heatmap classification and adopt L1 loss as $L_{reg}$ for box regression. $\lambda_{cls}$ and $\lambda_{reg}$ are weights for the two losses, respectively.

\section{EXPERIMENTS}
\label{sec:experiment}

\subsection{Experimental Settings}
\noindent
\textbf{Datasets.}
We conduct experiments on two large-scale datasets to evaluate proposed MMF-Track: KITTI tracking dataset~\cite{KITTI} and NuScenes dataset~\cite{nuScenes}. For the KITTI tracking benchmark, we follow the default setting of previous works~\cite{P2B,BAT,PTT} which adopts the sequences 0-16 for training, 17-18 for validation and 19-20 for testing. Moreover, different from the KITTI~\cite{KITTI}, NuScenes~\cite{nuScenes} is a larger dataset which contains a total of 1,000 scenes with 20s duration in each. Equipped with the 32-beam LiDAR sensor, NuScenes dataset contains about 300,000 points in every frame and has 360-degree view annotations for various objects. 
Compared to KITTI dataset, NuScenes is more challenging for object tracking since it has more challenging scenes. Additionally, it has been officially divided into training, validation and testing scenes. Following the setting of~\cite{BAT,beyond}, we also adopt the official 700 train scenes to train and the 150 val scenes to test.

\noindent
\textbf{Evaluation Metrics.}
For KITTI and NuScenes datasets, we use One Pass Evaluation (OPE) as evaluation metric, including \textit{Success} and \textit{Precision}. Specially, the \textit{Success} measures the 3D IoU between the predicted box and ground truth, ranging from 0 to 1, while the \textit{Precision} measures the accuracy between the predicted target center and ground-truth center, ranging from 0 to 2 meters. Following the previous work, we calculate the average performance based on the number of frames in each category and call it \textit{F-Mean}. Besides, we also adopt \textit{C-Mean} as a mean metric, which is obtained according to the number of categories.

\begin{table}[t!]
\renewcommand\tabcolsep{2.3pt}
\renewcommand{\arraystretch}{1.3}
\begin{center}
\caption{Performance comparison on the KITTI dataset. L means LiDAR-Only methods. LC denotes LiDAR-Camera methods. M and S are motion-based and similarity-based paradigms.}\label{tab:kitti}
		\begin{tabular}{c|c|c|cccc|cc}
			\toprule[.05cm]
			& \multirow{2}{*}{Method} & \multirow{2}{*}{Modality} & Car   & Ped. & 
                Van & Cyc. &F-Mean & C-Mean\\
			& & &\textit{6,424} &\textit{6,088} &\textit{1,248} &\textit{308} 
                &\textit{14,068} &\textit{14,068} \\
			\hline
			\hline
                \rule{0pt}{8pt}
			\multirow{14}*{\rotatebox{90}{\textit{Success}}}
			& SC3D~\cite{SC3D} & \multirow{11}{*}{L+S} &41.3 & 18.2 &40.4 &41.5 &31.2 &35.4\\
			& P2B~\cite{P2B} & &56.2 &28.7 &40.8 &32.1 &42.4 &39.5\\
			& PTT~\cite{PTT} & &67.8 &44.9 &43.6 &37.2 &55.1 &48.4\\
			& BAT~\cite{BAT} & &60.5 &42.1 &52.4 &33.7 &51.2 &47.2 \\
			& LTTR~\cite{lttr} & &65.0 &33.2 &35.8 &66.2 &48.7 &50.1\\
			& V2B~\cite{v2b} & &70.5 &48.3 &50.1 &40.8 &58.4 &52.4\\
                & C2FT~\cite{C2FT} & & 67.0 & 48.6 & 53.4 & 38.0 & 57.2 &51.8\\
                & MLSET~\cite{MLSET} & & 69.7 &50.7 & 55.2 & 41.0 & 59.6 &54.2\\
			& PTTR~\cite{pttr} & &65.2 &50.9 &52.5 &65.1 &57.9 &58.4\\
                & SMAT~\cite{SMAT} & &71.9 &52.1 &41.4 &61.2 &60.4 &56.7\\
                & STNet~\cite{stnet} & &\underline{72.1} &49.9 &\underline{58.0} &\underline{73.5} &61.3 &63.4 \\ \cline{2-9}
                \rule{0pt}{8pt}
                & M$^2$-Track~\cite{beyond} & L+M & 65.5 & \underline{61.5} &53.8 &73.2 &\underline{62.9} &\underline{63.5}\\ \cline{2-9}
                \rule{0pt}{8pt}
                & F-Siamese~\cite{F-siamese} & \multirow{2}{*}{LC+S} &37.1 &16.3 &- &47.0 &- &-\\ 
			& \textbf{Ours} & &\textbf{73.9} &\textbf{61.7} &\textbf{59.3} 
                &\textbf{75.9} &\textbf{67.4} &\textbf{67.7}\\
			\hline
			\hline
                \rule{0pt}{8pt}
			\multirow{14}*{\rotatebox{90}{\textit{Precision}}}
			& SC3D~\cite{SC3D} & \multirow{11}{*}{L+S} &57.9 & 37.8 &47.0 &70.4 &48.5 &53.3 \\
			& P2B~\cite{P2B} & &72.8 &49.6 &48.4 &44.7 &60.0 &53.9 \\
			& PTT~\cite{PTT} & &81.8 &72.0 &52.5 &47.3 &74.2 &63.4 \\
			& BAT~\cite{BAT} & &77.7 &70.1 &67.0 &45.4 &72.8 &65.1 \\
			& LTTR~\cite{lttr} & &77.1 &56.8 &45.6 &89.9 &65.8 &67.4 \\
			& V2B~\cite{v2b} & &81.3 &73.5 &58.0 &49.7 &75.2 &65.6 \\
                & C2FT~\cite{C2FT} & & 80.4 & 75.6 & 66.1 & 48.7 & 76.4 &67.7 \\
                & MLSET~\cite{MLSET} & & 81.0 & 80.0 & 64.8 & 49.7 & 78.4 &68.9 \\
			& PTTR~\cite{pttr} & &77.4 &81.6 &61.8 &90.5 &78.1 &77.8 \\
                & SMAT~\cite{SMAT} & &82.4 &81.5 &53.2 &87.3 &79.5 &76.1 \\
                & STNet~\cite{stnet} & &\underline{84.0} &77.2 &70.6 &\underline{93.7} &80.1 &81.4 \\ \cline{2-9}
                \rule{0pt}{8pt}
                & M$^2$-Track~\cite{beyond} & L+M &80.8 &\underline{88.2} &\underline{70.7} &93.5 &\underline{83.4} &\underline{83.3} \\ 
                \cline{2-9}
                \rule{0pt}{8pt}
                & F-Siamese~\cite{F-siamese} & \multirow{2}{*}{LC+S} &50.6 &32.3 &- &77.3 &- &-\\ 
			& \textbf{Ours} & &\textbf{84.1} &\textbf{89.3} &\textbf{72.5} 
                &\textbf{95.0} &\textbf{85.6} &\textbf{85.2} \\
			\toprule[.05cm]
		\end{tabular}
	\end{center}
\vspace{-0.1in}
\end{table}

\noindent
\textbf{Implementation Details.}
In data processing, due to the different sizes of the categories in datasets, we apply the different settings for point cloud voxelization in the raw point branch. Specially, for Car, we set point cloud range as [-3.2$m$, 3.2$m$] along the x-y axis, and voxel size is [0.025$m$, 0.025$m$, 0.05$m$]. For Van, we set the point cloud range as [-4.5$m$, 4.5$m$] and the voxel size as [0.0375$m$, 0.0375$m$, 0.05$m$]. And for Cyclist and Pedestrian, we set a smaller range and voxel size for their small size, which is [-1.2$m$, 1.2$m$] and [-1.8$m$, 1.8$m$] respectively, and voxel size is the same as Car. Meanwhile, all height ranges are [-3$m$, 1$m$] in KITTI but [-5$m$, 3$m$] in NuScenes dataset along the z-axis. The template and search points are voxelized following \cite{VoxelNet}. Then, a maximum of five points are randomly sampled from each voxel. For pseudo point branch, we process the number of points to $N=1024$ through randomly copying and discarding. Then, we adopt PointNet++~\cite{PointNet++} consisting of three set-abstraction (SA) layers to encode pseudo points. The ball-query radii are set to 0.3$m$, 0.5$m$ and 0.7$m$. We train 40 epochs for MMF-Track with Adam optimizer and set batch size 32 on NVIDIA 2080Ti GPU.

\begin{table}[t]
\renewcommand\tabcolsep{3.0pt}
\renewcommand{\arraystretch}{1.2}
\begin{center}
\caption{Performance comparison between ours and M$^2$-Track on the modified KITTI. Ped and Cyc is an abbreviation for pedestrian and cyclist.}~\label{tab:modified kitti}
\begin{tabular}{c|c|cccc|cc}
\toprule[.05cm]
\multirow{2}{*}{}          & Category     & Car  & Ped. & Van  & Cyc. & F-Mean & C-Mean\\
                           & Frame Number & \textit{1,328} & \textit{1,248}  & \textit{255}  & \textit{65}  & \textit{2,896} & \textit{2,896}\\ \hline \hline
\multirow{2}{*}{\textit{Success}}   & M$^2$-Track*  & 40.4 & 19.9   & 16.4 & 16.6 & 28.9 & 23.3\\
                           & \textbf{Ours}  & \textbf{43.0}  & \textbf{30.4} & \textbf{18.9}     & \textbf{21.0}  & \textbf{35.0} & \textbf{28.3}    \\ \hline\hline
\multirow{2}{*}{\textit{Precision}} & M$^2$-Track*  & 46.9 & 34.0   & 16.0 & 17.3    & 38.0 & 28.6\\
                           & \textbf{Ours}  & \textbf{47.2}  & \textbf{45.7} & \textbf{20.2}     & \textbf{21.5}  & \textbf{43.6} & \textbf{33.7}   \\ 
\toprule[.05cm]
\end{tabular}
\end{center}
\vspace{-0.25in}
\end{table}

\noindent
\textbf{Data Augmentation.}
To get the template and search data in the training phase, we first acquire the search point clouds and image by the index $X$. Then, we randomly select another index ranging $[X-5, X+5]$ from the same tracking sequence of the search data to obtain the template point clouds and image. In this way, the difference in target appearance between the template and search points will be randomly enlarged during training, which is beneficial for the network to better adapt to target appearance change and improve model generalization in the testing phase.

\begin{figure}[t]
\centering
\includegraphics[width=8.5cm,height=8.8cm]{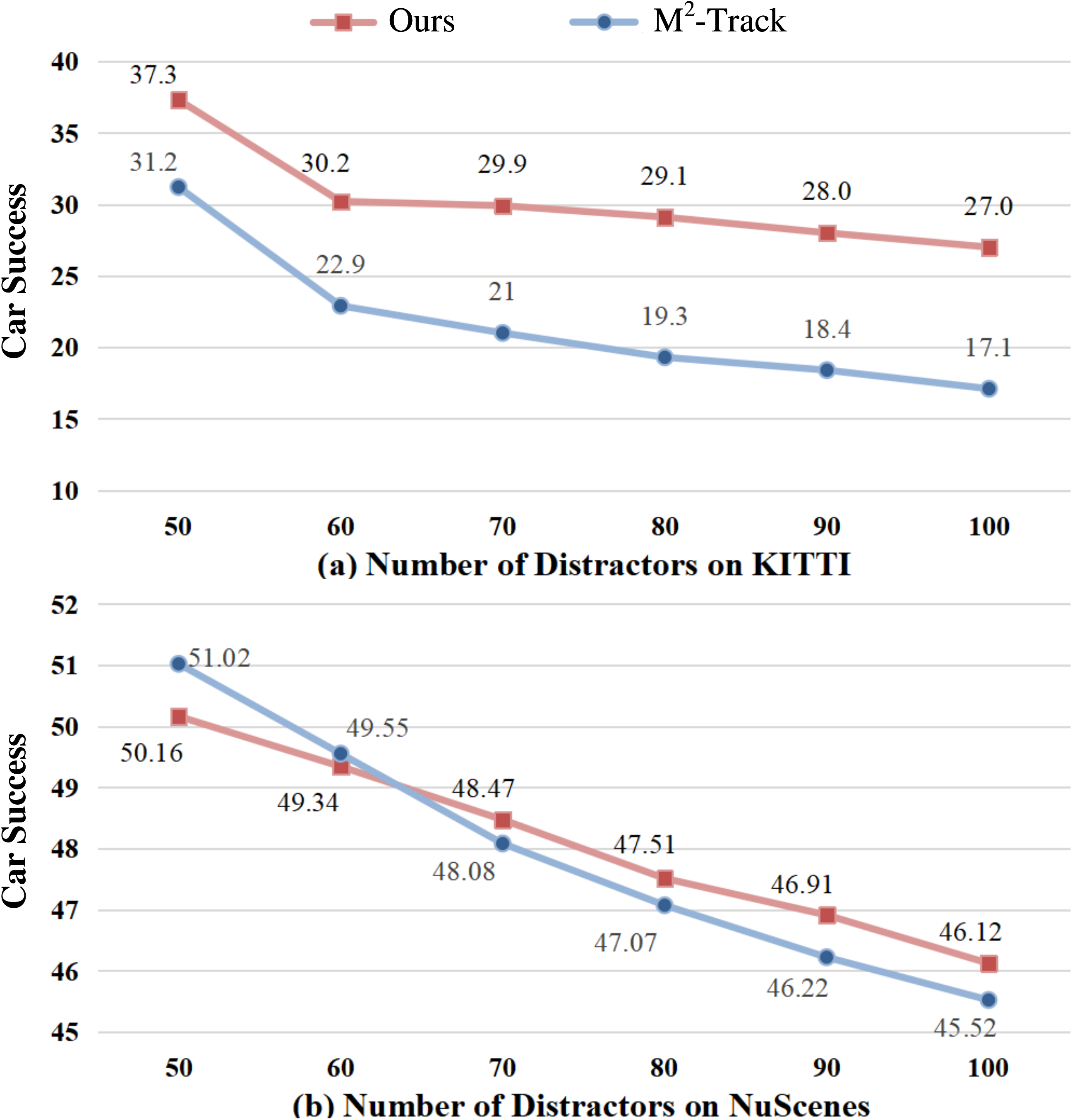} 
\caption{\textbf{Effect of the number of distractors on the tracking performance.} In (a) and (b), we compare the robustness of M$^2$-Track and our method in KITTI and Nuscenes respectively}
\label{fig:DB}
\vspace{-0.06in}
\end{figure}

\subsection{State-of-the-arts Comparison}
\noindent
\textbf{Quantitative results on KITTI.}
As demonstrated in Tab.~\ref{tab:kitti}, the former LiDAR-Only methods could be separated into the similarity-based~\cite{SC3D,P2B,PTT,BAT,lttr,v2b,C2FT,MLSET,pttr,SMAT,stnet}, and the motion-based~\cite{beyond} methods. Due to only relying on geometry matching of sparse points, the similarity-based paradigm has some limitations as mentioned in Sec.~\ref{sec:introduction}. In contrast, motion-based paradigm~\cite{beyond} considers target motion between two frames and achieves better performance than the matching-based methods. However, benefiting from introducing image information and texture matching, our multi-modal method outperforms \cite{beyond} by $4.5\%$ and $4.2\%$ on the \textit{F-Mean} and \textit{C-Mean} \textit{Success}. This result shows that utilizing multi-sensor data could develop the potential of similarity-based paradigm and surpass the motion-based method~\cite{beyond}.
Moreover, compared with LiDAR-Camera method F-Siamese~\cite{F-siamese}, we achieve a significant performance improvement in all categories. It validates that by multi-modal interaction and geometry-texture similarity fusion, our dual-stream structure could effectively integrate target clues from different modalities and is better than cascaded multi-modal tracker~\cite{F-siamese}. 

To test the tracking performance when the target is moving rapidly, we take one frame out of every 5 frames as the valid frame on KITTI. In this way, the relative motion between two frames will become bigger, which causes a greater challenge to the tracker. Nonetheless, in the modified KITTI, our method still outperforms M$^2$-Track~\cite{beyond} by 6.1\% and 4.4\% in \textit{F-Mean} performance in Tab.~\ref{tab:modified kitti}, which verifies that MMF-Track could adapt to large motion of the target. Moreover, similar to~\cite{beyond}, we randomly add car instances to the testing scenes of KITTI and test the robustness of tracker to distractors. As illustrated in Fig.~\ref{fig:DB}(a), when the number of distractors increased from 50 to 100, our method can maintain higher performance compared with M$^2$-Track~\cite{beyond}.

\begin{table}[t]
\renewcommand\tabcolsep{0.5mm}
\renewcommand{\arraystretch}{1.3}
\begin{center}
\caption{Performance comparison on the NuScenes dataset. * represents the results we tested with the official code.}~\label{tab:nuscenes multi-modal}
\begin{tabular}{l|c|cccc|cc}
\toprule[.05cm]
\multirow{2}*{}
& \multirow{2}{*}{Method} & Car & Ped. & Bus & Bic. & F-Mean & C-Mean\\
& &\textit{64,159} &\textit{33,227} &\textit{2,953} &\textit{2,292} &\textit{102,631} &\textit{102,631}\\
\hline
\hline
\rule{0pt}{8pt}
\multirow{2}*{\textit{Success}}
& TransFusion*~\cite{transfusion} & 37.42 & 12.85 & 32.44 & 19.45 & 28.92 &25.54\\
& \textbf{Ours} & \textbf{50.73} & \textbf{32.80} & \textbf{52.73} & \textbf{37.53} & \textbf{44.69} & \textbf{43.45}\\
\hline
\hline
\rule{0pt}{8pt}
\multirow{2}*{\textit{Precision}}
& TransFusion*~\cite{transfusion} & 46.52 & 26.82 & 36.01 & 41.96 & 39.74 &37.83\\
& \textbf{Ours} & \textbf{58.51} & \textbf{66.25} & \textbf{52.78} & \textbf{68.59} & \textbf{61.08} & \textbf{61.53}\\
\toprule[.05cm]
\end{tabular}
\end{center}
\vspace{-0.1in}
\end{table}

\begin{figure}[t!]
\centering
\includegraphics[width=7.8cm,height=4.8cm]{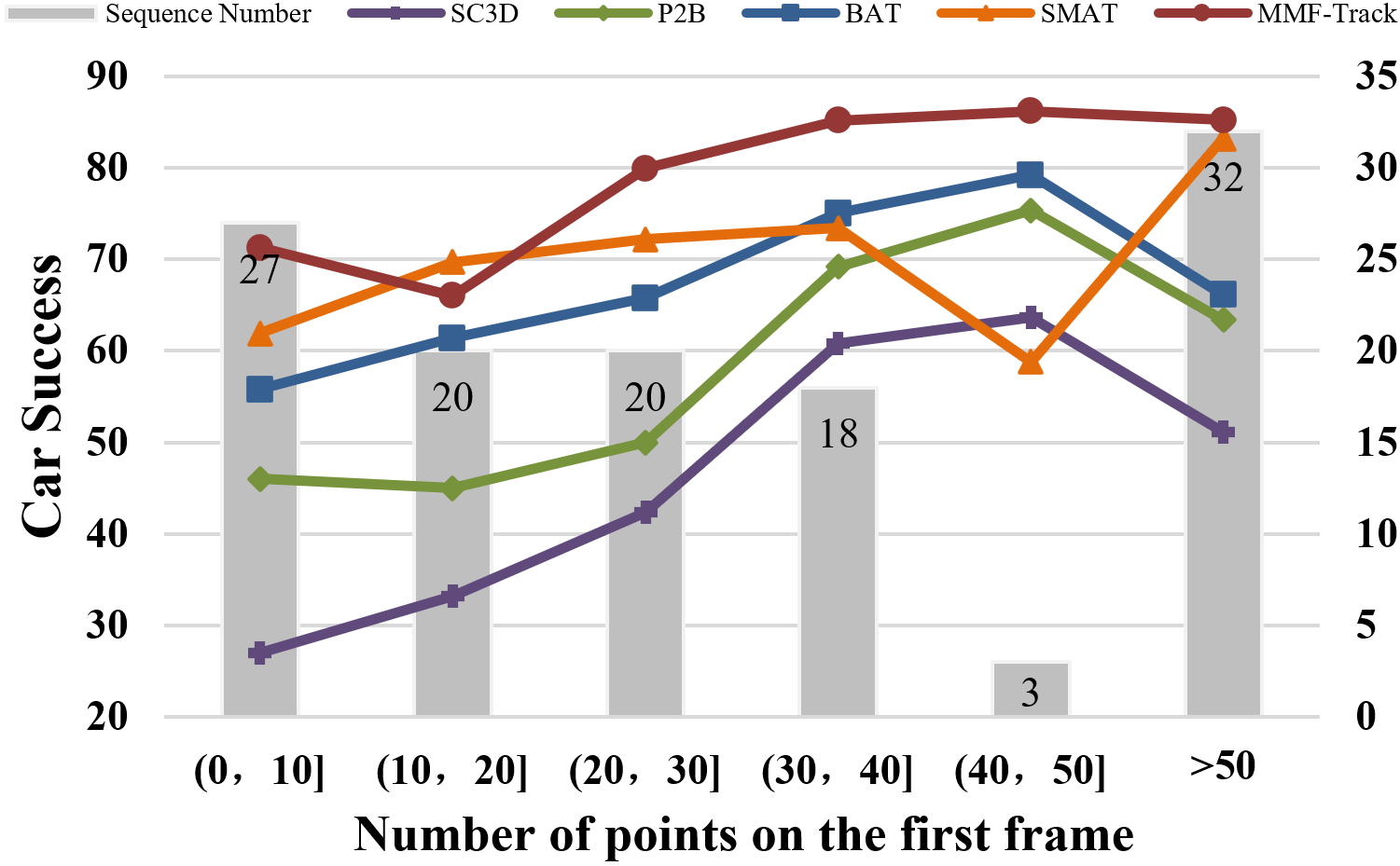} 
\caption{\textbf{Effect of the number of points in the first frame.} In most cases, our MMF-Track outperforms other methods by a significant margin.}
\label{fig:Number_Point}
\vspace{-0.1in}
\end{figure}

\noindent
\textbf{Quantitative results on NuScenes.}
Since there is no multi-modal SOT approach evaluated on NuScenes, we utilize the advanced multi-modal detection method, TransFusion~\cite{transfusion}, to perform multi-object tracking and transfer its results into the SOT task. As displayed in Tab.~\ref{tab:nuscenes multi-modal}, our method outperforms \cite{transfusion}, which may be attributed to the difficulty in using target-specific features for TransFusion.
Besides, MMF-Track obtains the best results among similarity-based methods~\cite{SC3D, P2B, BAT, PTT, SMAT} in Tab.~\ref{tab:nuscenes}, proving that our method can improve tracking performance by effectively combining geometry and texture similarities. Then, compared with the motion-based method~\cite{beyond}, our approach has advantages for tracking small-size objects. Especially for the Pedestrians, MMF-Track achieves $66.25\%$ in the \textit{Precision} and outperforms~\cite{beyond} by $5.33\%$. 
Furthermore, consistent with KITTI, we also validate the tracker after adding the different number of distractors in Fig.~\ref{fig:DB}(b). The results are shown that with the increase of distractors, the performance of M$^2$-Track will drop faster and even lower than our method. In contrast, MMF-Track exhibits more robust performance, which may be because image texture can help tracker to differentiate target from distractors and improve the robustness of tracker.

\begin{table}[t]
\renewcommand\tabcolsep{1.6pt}
\renewcommand{\arraystretch}{1.4}
\begin{center}
\caption{Performance comparison on the NuScenes dataset. Ped and Bic is an abbreviation for pedestrian and bicycle.}~\label{tab:nuscenes}
\begin{tabular}{l|c|c|cccc|cc}
\toprule[.05cm]
\multirow{2}*{}
& \multirow{2}{*}{Method} & \multirow{2}{*}{Modality} & Car & Ped. & Bus & Bic. & F-Mean & C-Mean \\
& & &\textit{64,159} &\textit{33,227} &\textit{2,953} &\textit{2,292} &\textit{102,631} & \textit{102,631}\\
\hline
\hline
\rule{0pt}{8pt}
\multirow{6}*{\rotatebox{90}{\textit{Success}}}
& SC3D~\cite{SC3D} & \multirow{5}{*}{L+S} & 22.31 & 11.29 & 29.35 & 16.70 & 18.82 & 19.91\\
& PTT~\cite{PTT} & & 41.22 & 19.33 & 43.86 & 28.39 & 33.94 & 33.45\\
& P2B~\cite{P2B} & & 38.81 & 28.39 & 32.95 & 26.32 & 34.99 & 31.62\\
& BAT~\cite{BAT} & & 40.73 & 28.83 & 35.44 & 27.17 & 36.42 & 33.04\\
& SMAT~\cite{SMAT} & & 43.51 & \underline{32.27} & 39.42 & 25.74 & 39.36 & 35.24\\ \cline{2-9}
\rule{0pt}{8pt}
& M$^2$-Track~\cite{beyond} & L+M & \textbf{55.85} & {32.10} & \underline{51.39}& \underline{36.32} & \textbf{47.60} & \textbf{43.92}\\
\cline{2-9}
\rule{0pt}{8pt}
& \textbf{Ours} & LC+S & \underline{50.73} & \textbf{32.80} & \textbf{52.73} & \textbf{37.53} & \underline{44.69}  & \underline{43.45}\\
\hline
\hline
\rule{0pt}{8pt}
\multirow{6}*{\rotatebox{90}{\textit{Precision}}}
& SC3D~\cite{SC3D} & \multirow{5}{*}{L+S} & 21.93 & 12.65 & 24.08 & 28.12 & 19.13 & 21.70\\
& PTT~\cite{PTT} & & 45.26 & 32.03 & 39.96 & 51.19 & 40.96 & 42.11\\
& P2B~\cite{P2B} & & 43.18 & 52.24 & 27.41 & 47.80 & 45.76 & 42.66\\
& BAT~\cite{BAT} & & 43.29 & 53.32 & 28.01 & 51.37 & 46.27 & 44.00\\
& SMAT~\cite{SMAT} & & 49.04 & 60.28 & 34.32 & 61.06 & 52.52 & 51.18\\
\cline{2-9}
\rule{0pt}{8pt}
& M$^2$-Track~\cite{beyond} & L+M & \textbf{65.09} & \underline{60.92} & \underline{51.44} & \underline{67.50} & \textbf{63.40} & \underline{61.24}\\
\cline{2-9}
\rule{0pt}{8pt}
& \textbf{Ours} & LC+S & \underline{58.51} & \textbf{66.25} & \textbf{52.78} & \textbf{68.59} & \underline{61.08} & \textbf{61.53}\\
\toprule[.05cm]
\end{tabular}
\end{center}
\vspace{-0.18in}
\end{table}

\begin{table}[t]
\renewcommand\tabcolsep{1.5mm}
\renewcommand{\arraystretch}{1.3}
\caption{Time consumption of each module in our MMF-Track.}~\label{tab:times}
\begin{tabular}{c|c|c|c}
\toprule[.05cm]
Module & Space Alignment       & Feature Encoder & Feature Interaction \\ \hline\hline
Time   & 14.4 $ms$             & 20.3 $ms$       & 25.1 $ms$           \\ \hline
Module & Similarity Generation & Regression Head & Overall             \\ \hline\hline
Time   & 13.2 $ms$             & 2.6 $ms$        & 75.6 $ms$           \\ \toprule[.05cm]
\end{tabular}
\vspace{-0.1in}
\end{table}

\begin{figure*}[t]
\centering
\includegraphics[width=\linewidth,height=15cm]{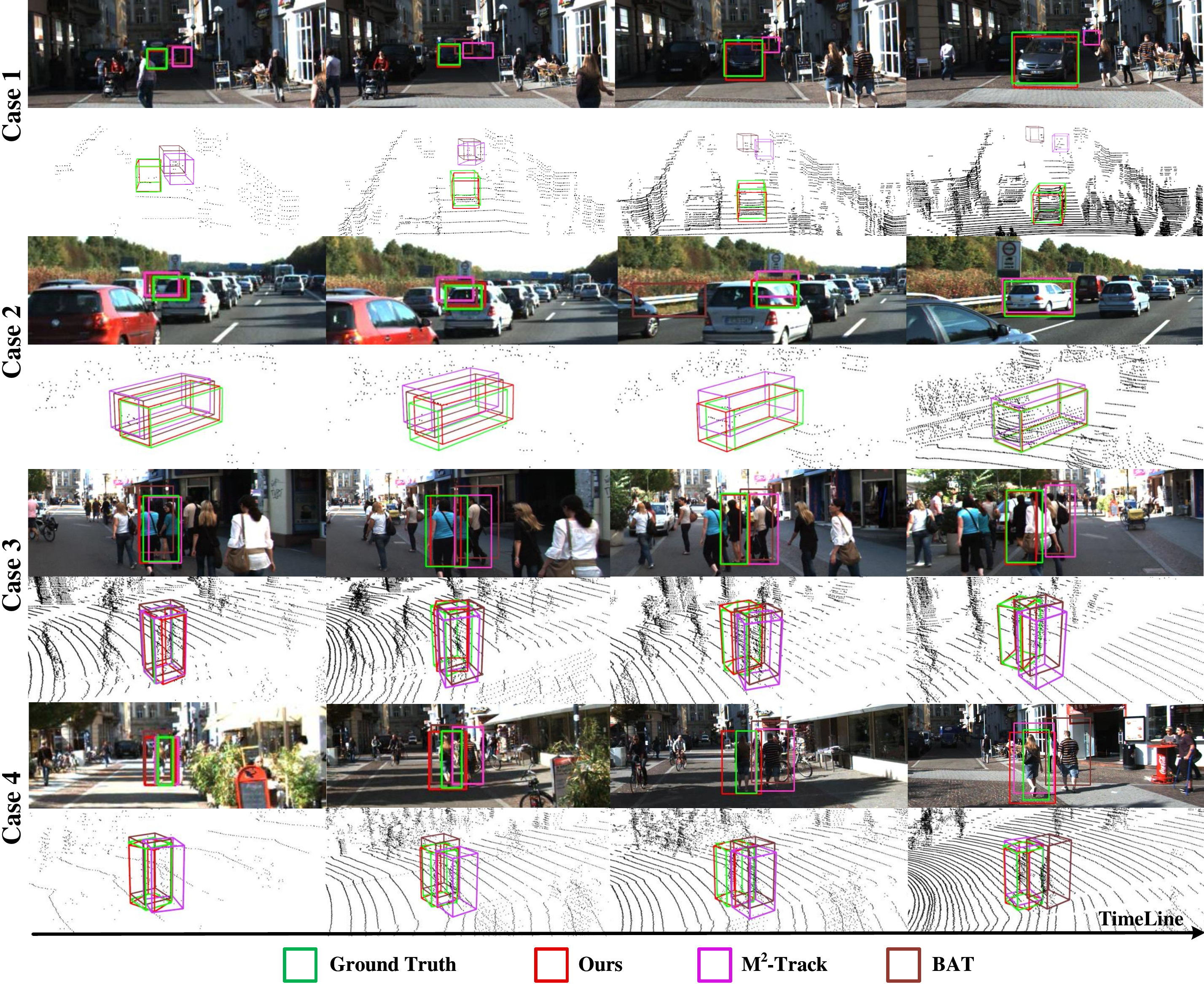}
\caption{\textbf{Visualization of tracking results on KITTI dataset.} From top to bottom, we compare our MMF-Track with BAT, M$^2$-Track and show the cases of Car and Pedestrian categories.}
\label{fig: Visulization}
\vspace{-0.05in}
\end{figure*}

\noindent
\textbf{Visualization results on KITTI.} We visualize tracking results of different categories in Fig.~\ref{fig: Visulization}. In the scene with a few points, such as the 1$^{st}$ and 2$^{nd}$ case, BAT and M$^2$-Track are prone to interference from backgrounds, resulting in the target loss. But our method can alleviate the impact of point cloud sparsity and better distinguish the target and backgrounds. When there are some objects similar to target (in the 3$^{rd}$ and 4$^{th}$ case), the M$^2$-Track and BAT both incorrectly capture similar objects due to depending on textureless points. In contrast, our MMF-Track achieves robust tracking results by using texture features.

\noindent
\textbf{Robustness to Sparseness.}
In Fig.~\ref{fig:Number_Point}, following~\cite{SC3D,P2B,BAT,SMAT}, we display tracking performance under the different number of points in the first frame. The result demonstrates that due to introducing rich texture information, MMF-Tracker could maintain high performance in most cases and achieve better robustness than other single-modal methods.

\noindent
\textbf{Running speed.} 
The time consumed by each module of the network is shown in detail in Tab.~\ref{tab:times}. We observe that MMF-Track can achieve 13.2 FPS on a single NVIDIA 2080Ti GPU and makes certain improvement compared with the 5 FPS of multi-modal F-Siamese~\cite{F-siamese}.

\subsection{Ablation Study}
We conduct ablation studies to validate the effectiveness of the proposed modules. Most experiments are performed on the Car category of KITTI.

\noindent
\textbf{Model Components.}
In Tab.~\ref{tab: ablation model components}, we demonstrate the effectiveness of our modules. Due to A1 only relying on geometry matching of sparse point clouds, its tracking result is not ideal. 
Then, compared with A5, removing any modules (in A2$\sim$A4) will result in performance degradation. Thus, every module is indispensable for the network. Specifically, A2 reflects that the template mask plays a critical role and helps tracker perceive target well by contextual background. Besides, as shown in A3 and A4, lacking multi-modal interaction or similarity fusion will also cause bad results, especially in A4, which validates that effectively using multi-modal information is the key to multi-modal tracker.

\noindent
\textbf{Feature Interaction.} We compare different feature interaction methods in Tab.~\ref{tab: feature interaction methods}. First, as shown in B1, although cross-attention is commonly utilized to fuse different modalities, it is not enough to construct complex multi-modal associations. Instead of simple two-in-one manner, FIM can enhance intra-modal features and construct inter-modal associations, which improves the \textit{Success}/\textit{Precision} by 1.8$\%$/2.0$\%$. Then, we can obtain further performance improvement by stacking $n=3$ FIMs. Lastly, benefiting from the coarse-to-fine strategy, B4 achieves the best performance, higher than B3 with 2.0\% in both \textit{Success} and \textit{Precision}, which proves that our CFIM has a stronger ability to construct semantic association.

\begin{table}[t]
\renewcommand\tabcolsep{1.5mm}
\renewcommand{\arraystretch}{1.3}
\centering
\caption{Ablation studies on model components. CFIM means Coarse-to-Fine Interaction Module. SFM represents Similarity Fusion Module. TM is the template mask.}
\begin{tabular}{c|c|c|c|c|c|c}
\toprule[.05cm]
{} & {Modality} & {TM} & {CFIM} & {SFM} & {3D Success} & {3D Precision} \\ \hline \hline
A1 & L & \XSolidBrush & \XSolidBrush & \XSolidBrush & 65.4 {\color{red}{$\downarrow$8.5}} & 74.6 {\color{red}{$\downarrow$9.5}}  \\
A2 & LC & \XSolidBrush & \Checkmark & \Checkmark & 70.4 {\color{red}{$\downarrow$3.5}} & 80.7 {\color{red}{$\downarrow$3.4}} \\
A3 & LC & \Checkmark & \XSolidBrush & \Checkmark & 70.6 {\color{red}{$\downarrow$3.3}} & 81.0 {\color{red}{$\downarrow$3.1}} \\ 
A4 & LC & \Checkmark & \Checkmark & \XSolidBrush & 70.0 {\color{red}{$\downarrow$3.9}} & 79.5 {\color{red}{$\downarrow$4.6}} \\ 
A5 & LC & \Checkmark & \Checkmark & \Checkmark & \textbf{73.9} & \textbf{84.1}  \\
\toprule[.05cm]
\end{tabular}
\label{tab: ablation model components}
\vspace{-0.1in}
\end{table}

\begin{table}[t]
\renewcommand\tabcolsep{1.0pt}
\renewcommand{\arraystretch}{1.3}
\centering
\caption{Ablation of the stream structure and feature interaction methods. FIM is the Feature Interaction Module.}
\begin{tabular}{c|c|c|c|c}
\toprule[.05cm]
{} & {Stream Structure} & {Interaction Strategy} & {3D Success} & {3D Precision} \\ \hline \hline
\rule{0pt}{8pt}
B1 & Two-in-One & Single Cross-Attention & 68.4 & 78.8 \\
B2 & Dual-Stream & Single FIM & 70.2 & 80.8 \\
B3 & Dual-Stream & Stacking $n$ FIMs & 71.9 & 82.1 \\
B4 & Dual-Stream & CFIM & \textbf{73.9} & \textbf{84.1} \\
\toprule[.05cm]
\end{tabular}
\label{tab: feature interaction methods}
\vspace{-0.1in}
\end{table}

\begin{table}[!t]
\renewcommand\tabcolsep{7pt}
\renewcommand{\arraystretch}{1.3}
\centering
\caption{Ablation studies of similarity fusion strategies.}
\begin{tabular}{c|c|c|c}
\toprule[.05cm]
{} & {Similarity Fusion Strategy} & {3D Success} & {3D Precision} \\ \hline \hline
\rule{0pt}{8pt}
C1 & Addition & 71.4 & 80.7\\
C2 & Concat & 71.3 & 81.8 \\
C3 & Cross-Attention & 70.6 & 80.3 \\
C4 & SFM & \textbf{73.9} & \textbf{84.1} \\
\toprule[.05cm]
\end{tabular}
\label{tab: similarity fusion method}
\vspace{-0.1in}
\end{table}

\begin{table}[!t]
\renewcommand\tabcolsep{10pt}
\renewcommand{\arraystretch}{1.3}
\centering
\caption{Ablation studies of different painting strategies.}
\begin{tabular}{c|c|c|c}
\toprule[.05cm]
{} & {Painting Strategy} & {3D Success} & {3D Precision} \\ \hline \hline
\rule{0pt}{8pt}
D1 & w/o Painting & 67.9 & 77.1 \\
D2 & w/ RGB & 71.5 & 81.2\\
D3 & w/ Feature Offline & 72.0 & 82.3\\
D4 & w/ Feature Online & \textbf{73.9} & \textbf{84.1} \\
\toprule[.05cm]
\end{tabular}
\label{tab: painting method}
\vspace{-0.05in}
\end{table}

\noindent
\textbf{Similarity Fusion.} As shown in
Tab.~\ref{tab: similarity fusion method}, we compare different similarity fusion strategies. Our proposed SFM achieves the best performance in C4, surpassing C1 by 2.5\% and 3.4\% in \textit{Success} and \textit{Precision}, respectively. Then, compared with directly using cross-attention in C3, our method obtains 3.3\% and 3.8\% improvement, showing the effectiveness of APM. It is interesting that the performance of C3 is lower than C1 and C2. We think that it's because directly using cross-attention may break the similarity between template and search features. But thanks to our APM that adaptively adjusts each-modal similarity in advance, we can effectively aggerate geometry-texture clues in the subsequent step.

\begin{table}[!t]
\renewcommand\tabcolsep{1.2pt}
\renewcommand{\arraystretch}{1.3}
\centering
\caption{The results of different template generation strategies.}
\begin{tabular}{c|c|c|c|c}
\toprule[.05cm]
Strategy                          & PTT~\cite{PTT}       & V2B~\cite{v2b}       & SMAT~\cite{SMAT}      & \textbf{MMF-Track}   \\ \hline \hline
\rule{0pt}{8pt}
First GT                    & 62.9/76.5 & 67.8/79.3 & 68.1/77.5 & \textbf{69.2}/\textbf{79.8}   \\
Previous result                 & 64.9/77.5 & 70.0/81.3 & 66.7/76.1 & \textbf{71.9}/\textbf{81.9}  \\
First GT \& Previous result & 67.8/81.8 & 70.5/81.3 & 71.9/82.4 & \textbf{73.9}/\textbf{84.1}   \\
All previous results            & 59.8/74.5 & 69.8/81.2 & 69.9/79.8 & \textbf{71.7}/\textbf{81.4}           \\
\toprule[.05cm]
\end{tabular}
\label{tab: template generation}
\vspace{-0.1in}
\end{table}

\begin{table}[t]
\renewcommand\tabcolsep{0.5pt}
\renewcommand{\arraystretch}{1.3}
\centering
\caption{Comparison tracking performance w/ or w/o image branch on the NuScenes.}
\begin{tabular}{c|c|c|c|c|c}
\toprule[.05cm]
\multirow{2}{*}{Image Branch} & Car & Ped.  & Bus & Bic. & F-Mean \\ 
&\textit{64,159} &\textit{33,227} &\textit{2,953} &\textit{2,292} &\textit{102,631}\\
\hline \hline
\Checkmark & \textbf{50.73}/\textbf{58.51} & \textbf{32.80}/\textbf{66.25} & \textbf{52.73}/\textbf{52.78}  & \textbf{37.53}/\textbf{68.59} & \textbf{44.69}/\textbf{61.08}  \\
\XSolidBrush  & 43.67/51.09 & 27.46/61.31 & 47.23/47.52  & 31.80/62.44 & 38.26/54.55 \\ \hline
\rule{0pt}{8pt}
\textit{Decreasing} & \color{Red}{7.06}/\color{Red}{7.42} & \color{Red}{5.34}/\color{Red}{4.94} & \color{Red}{5.50}/\color{Red}{5.26}  & \color{Red}{5.73}/\color{Red}{6.15} & \color{Red}{6.43}/\color{Red}{6.53}  \\
\toprule[.05cm]
\end{tabular}
\label{tab: ablation nus}
\vspace{-0.05in}
\end{table}

\noindent
\textbf{Painting Strategy.} In Tab.~\ref{tab: painting method}, we compare different painting strategies for pseudo points. Compared with not painting any information in D1, appending RGB to pseudo points achieves better results in D2 (3.6\%/4.1\% improvement). Furthermore, following~\cite{pointaugmenting}, we utilize semantic features extracted from an offline network to paint pseudo points in D3 and outperform D2 by 0.5\%/1.1\%. But the offline network can not be trained with tracker and will limit tracking performance. In contrast, in D4, our SAM can realize online painting and end-to-end training for the whole network, resulting in the best results.

\begin{figure}[!t]
\centering
\includegraphics[width=\linewidth,height=4.8cm]{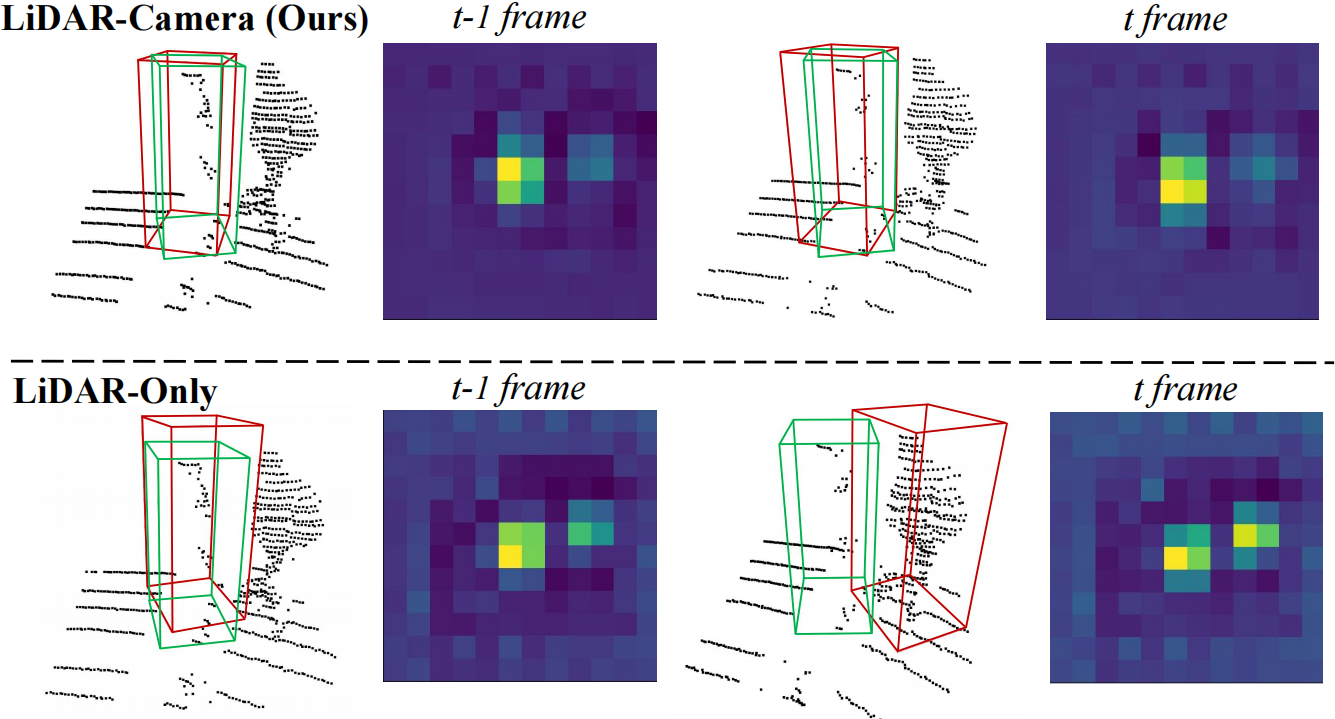} 
\caption{\textbf{Visual comparison of tracking results and heatmaps from two consecutive frames.} The predicted and ground-truth boxes are in red and green, respectively. }
\label{fig:Heatmap}
\vspace{-0.05in}
\end{figure}

\noindent
\textbf{Template Generation Strategy.}
As demonstrated in Tab.~\ref{tab: template generation}, we compare our MMF-Track with the previous works~\cite{PTT,v2b,SMAT} under different template generation strategies. The “first GT” means to adopt the ground-truth target points in the first frame as a template, while the “previous result” denotes using tracking result in the previous frame. The results show that our method achieves the best and relatively stable results on four template generation schemes, which proves the superiority of our multi-modal method.

\noindent
\textbf{LiDAR-Camera vs. LiDAR-Only.} To further verify the role of image in tracking, we first validate our MMF-Track without image branch on NuScenes. As shown in Tab.~\ref{tab: ablation nus}, the tracker occurs a certain performance degradation without texture feature. Then, for Pedestrian, we visualize tracking results of two consecutive frames and the corresponding heatmap in Fig.~\ref{fig:Heatmap}. The results display that the texture features extracted from the image could help tracker differentiate the target correctly when existing a distractor of the same class. Thus, we believe that image plays an important role in our multi-modal tracker.

\begin{figure*}[t]
\centering
\includegraphics[width=\linewidth,height=8.3cm]{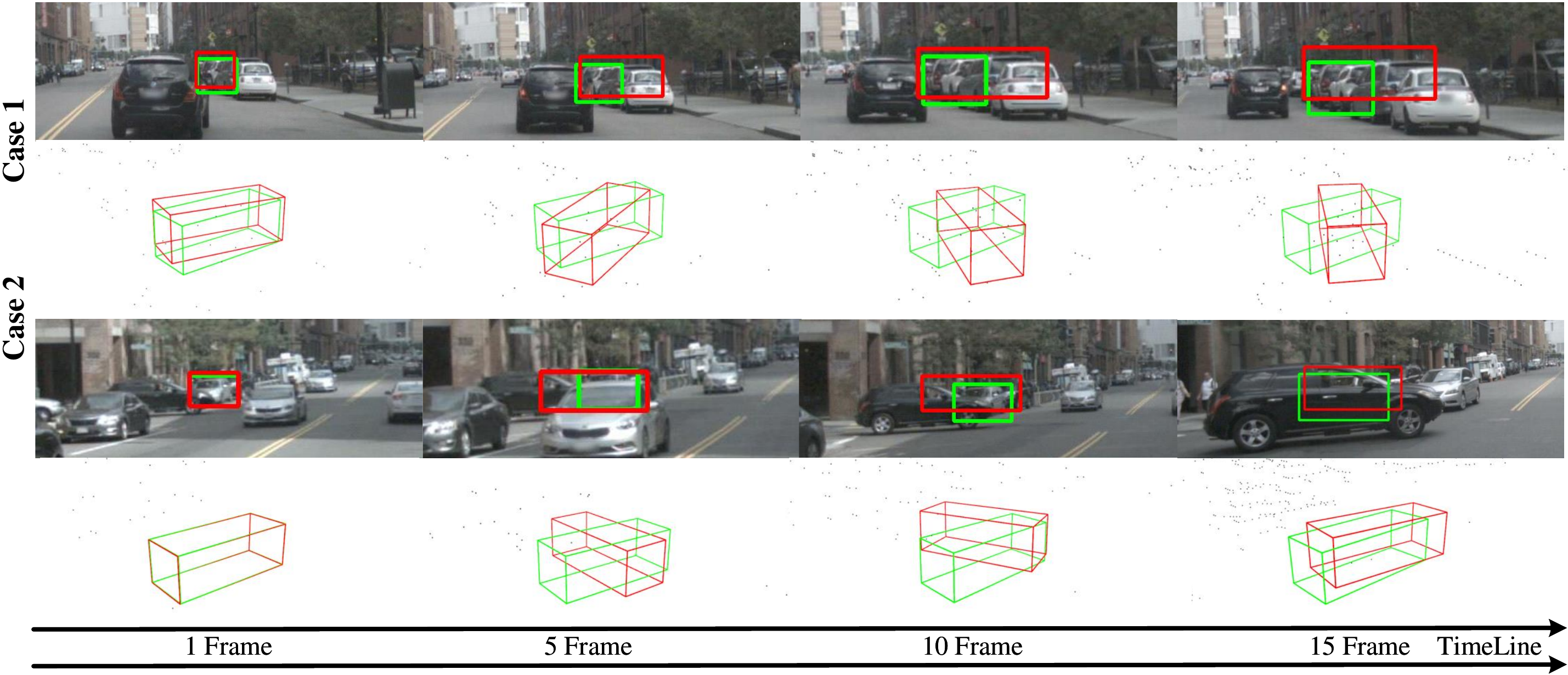} 
\caption{\textbf{Failures cases of our MMF-Track on the Nuscenes dataset.} The predicted and ground-truth boxes are displayed in red and green, respectively. }
\label{fig:Failure}
\vspace{-0.05in}
\end{figure*}

\subsection{Failure Cases}
\label{subsec:failure}
In Fig.~\ref{fig:Failure}, we illustrate some failure cases of MMF-Track on the Nuscenes, which has lower point density compared to KITTI. And we can observe when the target is occluded, the point clouds and pixels are rare or even missing, which makes it challenging for our MMF-Track to capture effective target characteristics and cause tracking failure. Therefore, in future work, we will further attempt to efficiently exploit more point cloud frames and extract the spatial-temporal information to improve MMF-Track. Besides, another improvement strategy is to integrate the long-term motion estimation with our multi-modal network through a more delicate design.

\section{Conclusion} 
\label{sec:conclusion}
In this paper, we introduce a novel multi-modal network MMF-Track for 3D SOT task. First, we design Space Alignment Module to generate textured pseudo points aligned with the raw points in 3D Space. Then, we propose a novel dual-stream structure to enhance intra-modal features and construct inter-modal semantic associations. Finally, we present a Similarity Fusion Module to aggregate geometry and texture clues from the target. Extensive experiments show that MMF-Track significantly outperforms previous multi-modal methods and achieves competitive results. We hope our work would inspire researchers to further study multi-modal 3D SOT.

\bibliographystyle{ieeetr}
\bibliography{my_egbib}

\end{document}